# Framework for Dynamic Evaluation of Muscle Fatigue in Manual Handling Work

Liang MA, Fouad BENNIS, Damien CHABLAT
{liang.ma, fouad.bennis, damien.chablat}@irccyn.ec-nantes.fr
Institut de Recherche en Communications et Cybernétique de Nantes, UMR CNR 6597
École Centrale de Nantes, 1, rue de la Noë
BP 92101, 44321 Nantes Cedex 03, France

Wei ZHANG
{zhangwei}@mail.tsinghua.edu.cn
Department of Industrial Engineering, Tsinghua University
Beijing, 100084, P.R.China

*Abstract* - Muscle fatigue is defined as the point at which the muscle is no longer able to sustain the required force or work output level. The overexertion of muscle force and muscle fatigue can induce acute pain and chronic pain in human body. When muscle fatigue is accumulated, the functional disability can be resulted as musculoskeletal disorders (MSD). There are several posture exposure analysis methods useful for rating the MSD risks, but they are mainly based on static postures. Even in some fatigue evaluation methods, muscle fatigue evaluation is only available for static postures, but not suitable for dynamic working process. Meanwhile, some existing muscle fatigue models based on physiological models cannot be easily used in industrial ergonomic evaluations. The external dynamic load is definitely the most important factor resulting muscle fatigue, thus we propose a new fatigue model under a framework for evaluating fatigue in dynamic working processes. Under this framework, virtual reality system is taken to generate virtual working environment, which can be interacted with the work with haptic interfaces and optical motion capture system. The motion information and load information are collected and further processed to evaluate the overall work load of the worker based on dynamic muscle fatigue models and other work evaluation criterions and to give new information to characterize the penibility of the task in design process.

## I. INTRODUCTION

For the workers performing the mechanical work, musculoskeletal disorder (**MSD**) is still a serious healthy problem. According to the report of HSE [1] and 1993 WA State Fund [2] compensable claims, over 50% industrial workers have suffered from MSD. After analysis by Don D.Chaffin [3], overexertion of muscle force or frequent high muscle load is the main reason to cause muscle fatigue, and further more, it results in acute muscle fatigue, pain in muscles and the worst functional disability in muscles and other tissues of human body. Hence, to find an efficient method to evaluate and predict the MSD risks becomes an essential mission for ergonomists.

Several posture exposure techniques are available for evaluating MSD risks. For general posture analysis, **Posturegram**, Ovako Working Posture Analysing System (**OWAS**), posture targeting and Quick Exposure Check (**QEC**) were developed. Further more, some special tools are designed for specified part of human body. For example, Rapid Upper Limb Assessment (**RULA**) is designed for assessing the severity of postural loading and is particularly applicable for sedentary jobs. The similar systems include **HAMA, PLIBEL** and so on [4, 5]. Even just for lifting job, five prevailing tools (**NIOSH lifting index, ACGIH TLV, 3DSSPP, WA L&I, Snook**) were developed and they are compared in Ref. [6]. Most of them have been verified useful for assessing MSD risks, but there are still several limitations. In posture exposure techniques, static postures are mainly considered, and it is not suitable for dynamic work processes. These techniques are not accurate enough to localize MSD risks to certain muscles. These methods are proposed independently, and they are not uniformed so that in some cases they could not get the similar result even for evaluating the same task.

For objectively predicting muscle fatigue, several muscle fatigue models and fatigue index have been proposed in publications. In a series of publications [7-9], Wexler and his colleges have proposed a new muscle fatigue model based on $Ca^{2+}$ cross-bridge mechanism and stimulation experiments, but it is mainly based on physiological mechanism and it is too complex for ergonomics evaluation, and further more, there are only parameters available for quadriceps. Taku Komura et al. have employed a muscle fatigue model based on force-pH relationship in computer graphics [10,11], and this model is used to visualize the muscle fatigue during a working process, but it just evaluated the fatigue of muscle at a time instant and cannot evaluate the overall muscle fatigue of the working processes. In this pH model, the force generation capacity can be mathematically analyzed, but the influences of the external forces are not considered enough. Rodiriguez proposes a half-joint fatigue model [12-14], more exactly a fatigue index, based on mechanical properties of muscle groups. In this half-joint model, it cannot predict individual muscle fatigue due to its half-joint principle. The maximum endurance equation of this model was from static posture analysis and it is only suitable for evaluating static postures.

In conventional methods, the evaluation must be carried out in field environment. Even with camera and other video equipments, the video has to be further processed to analyze the posture of the worker, so it is time consuming. Thus, virtual reality techniques, especially virtual human (manikin) are used to evaluate the human performance. Jack [15] is integrated into SIEMENS system for ergonomic analysis; virtual solider program [16] has created Santo for human performance analysis. But in both systems, there is no fatigue index for human fatigue evaluation.

In fact, most of the manual-handling jobs are implemented in dynamic process. To evaluate the dynamic procedure of the work can give ergonomists much more detailed information and thus we can enhance the quality, security of the jobs and improve working conditions and decrease MSD risks. In this paper, we are going to propose a new fatigue index and a VR framework under which muscle fatigue under dynamic working process can be evaluated in virtual environment.

## II. DYNAMIC MUSCLE FATIGUE INDEX

Work-related MSD have been found to be associated with numerous occupational "risk factors", including physical work load factors such as force, posture, movement and vibration, psychosocial stressors and individual factors.

For static or quasi-static works, posture exposure analysis techniques can evaluate and predict MSD risks; body posture is the only input parameter of these analysis tools. For dynamic analysis, the motion of the worker plays the same role as posture in posture exposure analysis.

Force is another important input factor for our framework. Definitely the fatigue of muscle or the performance of the worker is directly related with the load on the human body. The intensity, repetitiveness and duration of the external load should be considered as the main factor to evaluate the fatigue and other aspects of human work.

Individual factors, like gender, age, and weight and so on should also be taken into consideration. This information can be achieved from anthropometrical database and several biomechanical databases. In our framework, psychosocial stressors are not considered.

TABLE 1. SYMBOLS USED IN DYNAMIC FATIGUE INDEX

| Symbol | Unit | Description |
|---|---|---|
| $U$ | min | Fatigue Index to evaluate the muscle fatigue, fatigue feeling |
| $MVC$ | N | Maximum voluntary contractile force, constant and determined by personal factors |
| $F_{cem}(t)$ | N | Current exertable maximum force of muscle at time instant $t$ |
| $F_{Load}(t)$ | N | External load of muscle at the time instant $t$ |
| $k$ | min$^{-1}$ | Constant Parameter |

In manual handling work, muscle fatigue is determined mainly by the external load and personal factors. In subjective fatigue evaluation, the fatigue is mainly evaluated based on the feeling of the tester and our fatigue index is going to evaluate the fatigue feeling of the worker. The general feeling of our human body about fatigue is: the larger the force is, the faster we can feel fatigue; the longer the force maintains, the more fatigue; the smaller capacity of the muscle is, the more easily we can feel fatigue. Based on this description, the differential equation of the feeling of fatigue can be written as:

$$\frac{dU}{dt} = \frac{MVC}{F_{cem}(t)} \cdot \frac{F_{Load}(t)}{F_{cem}(t)} \quad (1)$$

Meanwhile, the current maximum exertable force $F_{cem}$ is changing with the time due to the external loads. It is sensible that: The larger the external load, the faster $F_{cem}$ decreases; the smaller $F_{cem}$ is, the more slowly $F_{cem}$ decreases. The differential equation for the $F_{cem}$ is:

$$\frac{dF_{cem}(t)}{dt} = -k \frac{F_{cem}(t)}{MVC} \cdot F_{Load}(t) \quad (2)$$

The integration result of equation (2) is:

$$F_{cem}(t) = MVC \cdot e^{-\int_0^t k \frac{F_{Load}(u)}{MVC} du} \quad (3)$$

Assume that $F(t)$ is:

$$F(t) = \int_0^t \frac{F_{load}(u)}{MVC} du \quad (4)$$

The feeling of our fatigue is a function below and which is closed related to $MVC$ and $F_{load}(t)$. MVC represents the personal factors, and $F_{load}(t)$ is the force exerted on the muscle along the time and it reflects the influences of external loads.

$$U = \frac{1}{2k} e^{2kF(t)} - \frac{1}{2k} e^{2kF(0)} \quad (5)$$

In this model, personal factors and external load history are considered to evaluate the muscle fatigue. It can be easily used and integrated into simulation software for real time evaluation especially for dynamic working processes. This model still needs to be mathematically validated and ergonomic experimental validated, and meanwhile muscle recovery procedure should also be included to make the fatigue index completed.

## III. FRAMEWORK FOR OBJECTIVE WORK EVALUATION SYSTEM

### A. Function Structure Analysis of the Framework

The overall objective of the framework is to evaluate human work and predict potential human MSD risks dynamically, especially for human muscle fatigue. The function structure of the framework is shown in Fig. 1 and discussed below.

The overall function of our framework is to field-independently evaluate the difficulty of human mechanical work including fatigue, comfort and other aspects. Here we are focusing on fatigue evaluation.

In order to avoid field-dependent work evaluation, virtual reality techniques and virtual human techniques are used. Immersive work simulation system should be first constructed to provide the virtual working environment. Meanwhile, virtual human should be modelled and driven by the motion capture data to map the real working procedure into the virtual environment. Haptic interfaces can be used to enable the interactions between the worker and virtual environment.

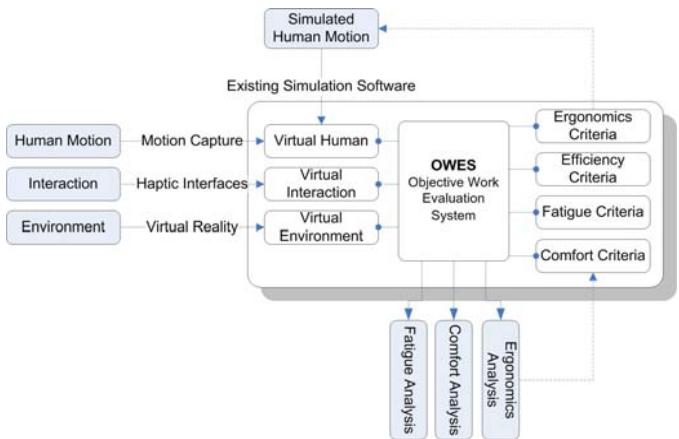

Fig. 1 Function structure of Framework of Objective Work Evaluation System

For any ergonomic analysis, data collection is the first important step. All the necessary information needs to be collected for further processing. From part 2, necessary information for dynamic manual handling jobs evaluation consists of motion, forces and personal factors. Motion capture techniques can be applied to achieve the motion information. In general, there are several kinds of tracking techniques available, like mechanical motion tracking, acoustic tracking, magnetic tracking, optical motion tracking and inertial motion tracking. Each tracking technique has its advantages and drawbacks for capturing the human motion. Hybrid motion tracking techniques can be taken to compensate the disadvantages and achieve the best motion data.

Force information can be recorded by haptic interfaces. Haptic interface is the channel via which the user can communicate with virtual objects through haptic interactions, and the interaction data between the worker and the virtual environment are also significant for evaluating other ergonomic aspects. Individual factors can be achieved from anthropometrical database and some biomechanical database.

All the information, such as motion information, force history and interaction events, is further processed into Objective Work Evaluation System (**OWAS**). The output of the framework is evaluation results of the mechanical work. There are many ergonomic aspects of a mechanical work. For each aspect, corresponding criteria should be established to assess the dynamic work process. Further more, these criteria can also be applied just into commercial human simulation software to generate much more naturally and realistically human simulation. Meanwhile, the output information can be used as feedback to improve these evaluation criteria.

In summary, all the necessary information describing the working process can be captured by peripheral devices like motion capture and haptic interfaces. Combined with virtual environment, the working information is further processed into OWAS with different criteria to evaluate different aspects of human manual handling work. Further more, evaluated result can be used to improve the evaluation criteria. With improved criteria, more natural human actions can be generated and simulated. At last, it is possible to simulate human working process as real as possible.

### B. Technical Specifications for the Framework

In this part, the technical requirements for three main function modules (Motion Tracking, Virtual Reality Simulation and Work Evaluation) are discussed.

**Motion Tracking**

Tracking module is used to digitalize the worker's operation and prepare the data for further processing. In order to record the worker's operation, tracking the worker's operation is the first task. In general, the technical requirements for the trackers are: tiny, self-contained, complete, accurate, fast, immune to occlusion, robust, tenacious, wireless and cheap. These are the requirements for the ideal tracker, but actually, every tracker available today falls short on at least seven of these 10 characteristics [17]. The performance requirements and purposes of the application are the decisive factors to select the suitable tracker.

In our framework, worker's operation needs to be tracked and digitalized, so that the positions of the worker's limbs should be known and as well the detailed motions- finger movements. The position of the worker's limbs determines the overall posture, and the motion of fingers represents the handling status of the task. In this case, several requirements should be fulfilled for this specific application.

- Tracking speed

Tracking worker's operation is easier than tracking athlete's performance, because generally there's no running or jumping in work. The tracker should satisfy tracking general movements of human body, and data update rate should be at least 24/25 Hz in order to realize real time visualization.

- Robustness

Worker's motion is tracked during performing certain tasks. During the working, the tracking should be stable and prevent influences from noises and other factors.

- Completeness

No tracker is suitable for tracking full-body motion and finger motion at the same time, integration of different trackers is necessary in order to capture all necessary motion information to describe worker's operation.

- Absolute accuracy

In general, applications demand accuracy with resolution 1mm in position and 0.1 degree in orientation. For full-body motion tracking, the demands are reduced in applications like character animation and biometrics. In this framework, the demand for accuracy depends on the types of the job. For general moves, the demands for accuracy are not very critical, but for some actions, like using tools or controlling switches, with interactions with virtual objects, the accuracy should be as high as possible.

- Data transferring

Transferring the data from tracking module to the other modules is another problem. Generally, there are real-time and no-real-time modes. In the latter manner, tracking data can be saved in file as data preparation. In real time manner, it is necessary to transfer the data to the simulation module as quickly as possible to ensure the real-time simulation.

**Virtual Reality Simulation**

General requirements for simulation are: feeling no sick during simulation, the visual content of the simulation should be updated in real time manner. In our framework, the simulation module is to simulate the worker's operation in the virtual working environment, provide visualization of the simulation to the worker performing the task and visualization the interactions between worker and virtual objects. It includes three parts: virtualization of virtual environment, visualization of virtual human and feedbacks between virtual simulation and worker.

- Visualization of virtual environment

Virtual working environment should be prepared, so that at least the operator can have the same spatial feelings as working in a field area. The virtual environment can be the copy of the real field environment or redesigned for new work environment validation. In the virtual environment there are fixed and movable virtual objects. Fixed objects can be work station, machine tools, working plats which remain stationary no matter how the user interacts with them; movable virtual objects can be for example some parts, bolts and boxes which can be moved in the simulation when the user moves them, changeable objects like buttons, switches which changes its state while the user interacts with them.

- Visualization of virtual human

Besides virtual working environment, virtual human should be modeled to present the worker's operation virtually. The virtual representation of the human is mapped into the virtual environment by the motion tracking data and can assist the worker working in the virtual environment and can give the observer the overview of the worker's operation. The virtual human should at least have the same dimension and appearance as the real human. This objective can be achieved by modeling human with anthropometrical database.

For muscle fatigue evaluation, it requires that the muscles need to be modeled into the virtual human. It also demands that the skeleton structure of human should also be modeled to determine the linkage relation between muscles and bones. After skeleton and muscle modeling, it is possible to compute the force of each individual muscle during the dynamic process exactly. For skeleton and muscle modeling, anatomical database can be used.

Virtual human can be modeled from anthropometrical and anatomical database, but it is still necessary to adjust the virtual human according to tracking data because tracking data is originally from an individual person, otherwise there will be errors in the mapping relationship between the worker and the virtual human.

- Interaction between the user and the virtual objects

In real-time simulation, interactive feedbacks should be provided in order to create immersive working feelings. Those interactive feedbacks include visual feedback, haptic feedback, and acoustic feedback and so on. The simulation module should have at least one view based on the viewpoint of the worker, so that correct visual feedback can be supplied to the worker.

Haptic feedback can give the user correct feelings of touching virtual objects, grasping or moving them. To ensure the fidelity of the haptic feedback, there is one critical requirement for the haptic feedback interfaces: high update rate (300Hz -1000Hz) [18]. In order to provide the right feedback, the interaction between worker and virtual objects should be real-time detected and analyzed. The coupling between the worker and the virtual objects should be simulated, for example, lifting a box. Besides that, feedback forces should be calculated with correct and efficient models to ensure the high update rate.

**Work Evaluation**

This part plays the role as ergonomists to evaluate the working process objectively. The evaluation criteria should be applied or designed into this framework to assess the work. There are several motion analysis techniques available for ergonomic and biomechanical analysis, and in our framework, these techniques should be re-designed suitable for computer analysis.

*C. A Prototype System of the Framework*

Based on the function analysis of the framework, a prototype system is under construction to realise dynamic fatigue evaluation.

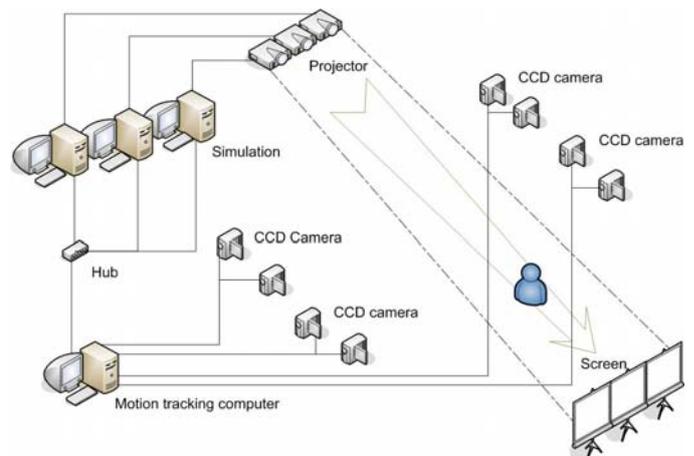

Fig. 2 hardware layout of the prototype system for dynamic work evaluation framework

Three hardware systems are employed to construct a prototype to realize the framework: virtual simulation system, motion capture system and haptic interfaces.

Virtual Simulation system consists of graphic simulation module and display module. Simulation module executes on a computer graphic station, and display module is composed of projection system and head mounted display (HMD). Simulation module is in charge of graphic processing and display control. The projection system and HMD system can display the immersive environment. Motion Capture system has been widely used in entertainment, 3D animation, navigation and so on. In this prototype, it is taken to capture the movement of the worker to collect the real working data. The Optical Motion Capture System developed by Tsinghua University VRLIB [19] is used, and it can work at 25 Hz data update rate. Each key joint of human body is attached by an active marker, and the overall skeletal structure of human body can be represented by 13 markers. Meanwhile, data gloves are used to capture the movement of fingers. Haptic system is also going to be integrated into this prototype, and now we are still in preparation stage.

Two software modules are developed to support these hardware systems, and they are simulation module and evaluation module. Based on Vega and MFC, the virtual simulation software can load the virtual environment CAD model to visualize the virtual working environment, and the virtual human composing of basic body parts can represent the worker in the virtual field environment. The dimension of the virtual human is fixed with 50% anthropometrical data. The data transferred from tracking software can drive the virtual worker performing the same action. In evaluation module, the different aspects of manual handling work can be assessed with different criteria. Haptic interaction and fatigue index is still under construction. The prototype is verified with motion data and MOST motion-time analysis to evaluate the work efficiency. Related figures of the prototype are shown below.

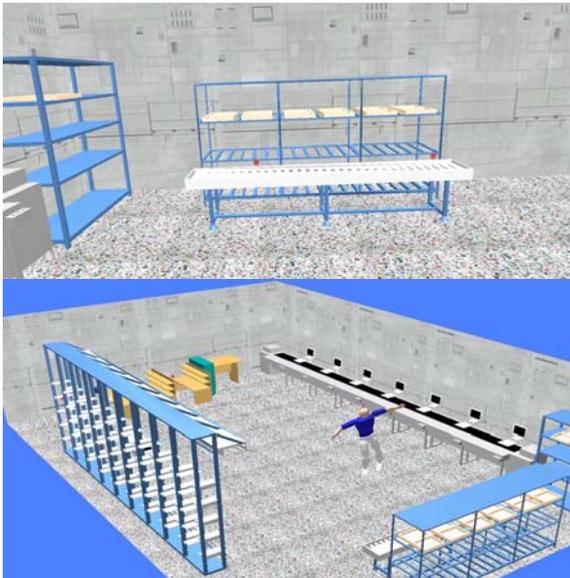

Fig. 3 Simulation results from different viewpoints. The upper one is based on the viewpoint of the worker and the lower one is the overall view.

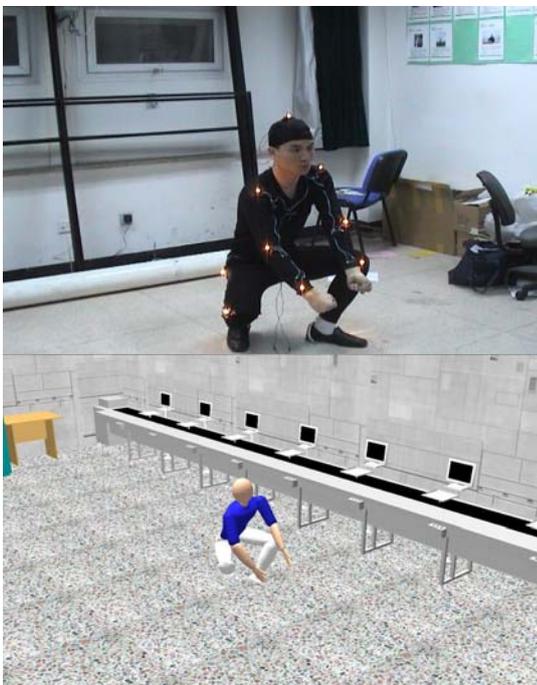

Fig. 4 Tracking result in the simulation software. The upper one shows the action of worker and the same action of the virtual human in the simulation software in the lower one.

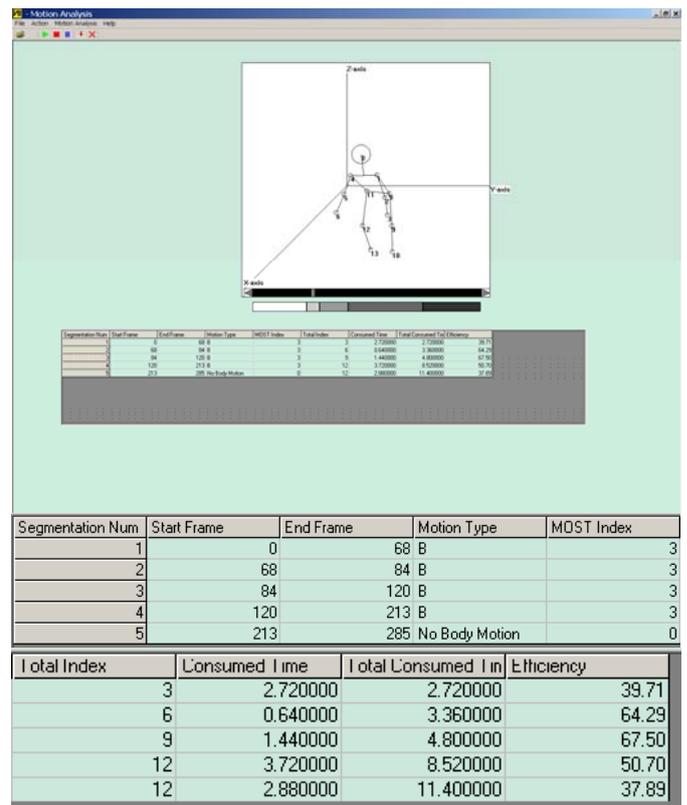

Fig. 5 Analysis result for efficiency evaluation. The analysis result is based on a lifting work. It segments the overall procedure into several parts which are compared to MOST standards to calculate the efficiency.

In Fig. 3, it shows different visual simulation result of the human motion, one from overview and the other one from worker's own view. Fig. 4 shows the original action of the worker and the action simulated with the tracking data in the software. Fig. 5 shows the evaluation analysis result based on Maynard Operation Sequence Technique (**MOST**) technique. A lifting procedure is analyzed based on MOST motion-time analysis technique. The results verify the structure of the framework and working flow of the objective evaluation system.

## IV. CONCLUSIONS AND PERSPECTS

In this paper, we presented a framework for dynamic human performance evaluation. Under this framework, it is possible to evaluate dynamic human handling work with the supports of virtual reality simulation, human motion tracking and haptic interfaces. A prototype system is now under construction in IRCCyN to assess human muscle fatigue of manual handling jobs.

The prototype system can evaluate not only the fatigue, but also the other aspects of work by extending the evaluation criteria and data collection modules. If the framework can be verified in the future, it could be integrated into other CAD systems to optimize the product design and work process design.

In future works, we will apply our results to enhance the simulation software such that they will able to produce realistic simulations.

## ACKNOWLEDGMENT

This research was supported by the EADS and by the Region des Pays de la Loire (France).